\documentclass[twoside,leqno,twocolumn]{article}
\usepackage{ltexpprt}
\usepackage{graphicx,subfigure,amsmath,multirow,xcolor,url}

\begin{document}

\title{\Large A Deep Spatio-Temporal Fuzzy Neural Network for Passenger Demand Prediction}
\author{Xiaoyuan Liang, Guiling Wang\thanks{\{xl367, gwang\}@njit.edu, New Jersey Institute of Technology, Newark, NJ. } \\
\and
Martin Renqiang Min\thanks{renqiang@nec-labs.com, NEC Laboratories America, Inc., Princeton, NJ.}\\
\and
Yi Qi\thanks{yi.qi@tsu.edu, Texas Southern University, Houston, TX.}\\
\and
Zhu Han\thanks{zhan2@uh.edu, University of Houston, Houston, TX. }}
\date{}

\maketitle


\fancyfoot[R]{\scriptsize{Copyright \textcopyright\ 2019 by SIAM\\
Unauthorized reproduction of this article is prohibited}}





\begin{abstract} \small\baselineskip=9pt 
In spite of its importance, passenger demand prediction is a highly challenging problem, because 
the demand is simultaneously influenced by the complex interactions among many spatial and 
temporal factors and other external factors such as weather. To address this problem, we 
propose a Spatio-TEmporal Fuzzy neural Network (STEF-Net) to accurately 
predict passenger demands incorporating the complex interactions of all known important factors. 
We design an end-to-end learning framework with different neural networks modeling different factors.
Specifically, we propose to capture spatio-temporal feature interactions via a 
convolutional long short-term memory network and model external factors via a fuzzy 
neural network that handles data uncertainty significantly better than deterministic methods. 
To keep the temporal relations when fusing two networks and emphasize discriminative spatio-temporal 
feature interactions, we employ a novel feature fusion method with a convolution operation and an attention 
layer. As far as we know, our work is the first to fuse a deep recurrent neural network and a fuzzy neural network to model complex 
spatial-temporal feature interactions with additional uncertain input features for predictive learning. Experiments on a 
large-scale real-world dataset show that our model achieves more than 10\% improvement over the state-of-the-art approaches.
\end{abstract}

\section{Introduction}
\label{sec:intro}

Accurate future passenger demand prediction is very important in the field of transportation. 
Knowing the future demands, a Transportation Network Company (TNC) can wisely pre-allocate resources (vehicles and drivers) 
to meet the demands, such that the best service can be provided to passengers with a minimum waiting time, 
and unnecessary driving around on road can be prevented, reducing energy consumption and traffic jam. 
However, passenger demand prediction is very challenging considering the future demands are 
simultaneously influenced by many factors, including continuous spatial and temporal factors, as well as many discrete 
external factors, such as weather and being daytime or nighttime. 
These factors have complex and non-linear interactions with future demands and capturing the interactions in one model to make prediction is very
difficult. Moreover, data of external factors are often either inaccurate 
or too coarse due to data-collection sensors' sparse deployment and unavoidable errors. 

To predict passenger demands in the near future, previous works have proposed various models. 
One of the most well-known methods uses the Auto-Regressive Integrated Moving Average (ARIMA) \cite{SheWil08_y,MorGam13_y}, 
but it only considers the temporal feature interaction.
Recent studies \cite{ZhoShe18,ZhaZhe17_y} propose deep learning models considering 
both temporal and spatial feature interactions, which outperform previous methods considering only one type of factor.
However, they handle the spatial and temporal feature interactions sequentially, resulting in information loss.
In addition, they ignore other important discrete external factors, e.g., the weather. 
Although the model proposed in \cite{KeZhe17} considers the weather impact, 
it fails to consider inaccuracies within the collected data. 
 
To tackle this challenging problem with desirable performance, 
we propose a deep Spatio-TEmporal Fuzzy neural Network (STEF-Net) to predict the passenger demands for a city area. 
In our network, we fuse all related factors to model the complex interactions among them, including spatial-temporal dependencies, 
external information and temporal relevance, and design an end-to-end learning framework with different neural networks 
modeling different types of feature interactions. Specifically, our model simultaneously captures the spatial and temporal 
dependencies via a Convolutional Long Short-Term Memory network (ConvLSTM). 
A ConvLSTM replaces the full connections in a traditional LSTM 
with convolutional operations such that the spatial feature interactions can be simultaneously captured and information loss can be avoided 
compared to sequential processing by stacking convolutional layers and LSTMs.
Regarding the uncertain accuracy of external factors, we propose to model them via a fuzzy neural network.
A fuzzy neural network, which combines the fuzzy theory and neural networks, can learn  
the feature representation with high error tolerance and trainable rules. 
It shows significantly better performance than deterministic neural networks for this data type. 
We design a new feature fusion method using a convolution operation to connect the two separate networks without losing temporal information. 
We further propose to capture the temporal relevance of the high-level fused data via an attention layer, 
considering the future demands are unequally influenced by past ones. 
Our model is evaluated on the real data from Didi Chuxing, the biggest TNC in China, which is similar to Uber in the United States. 
The experimental results show that our model outperforms the state-of-the-art models.

In summary, our contributions include: 
(1) To the best of our knowledge, our work is the first to combine a fuzzy neural network and deep learning techniques to handle data uncertainty and learn complex interactions among multiple factors,
which can achieve better performance than solely using deep learning.
(2) We propose a new feature fusion method using a convolution operation, which can preserve the temporal relations of the outputs, capture the spatial information and achieve better performance than commonly used weighted addition.
(3)  We adopt an attention layer on every time step to provide explainable results on when the historical information influences most in the prediction and how the weather can influence the prediction.
(4) Extensive experiments are conducted on real data to evaluate our model.
Our model significantly outperforms state-of-the-art models in prediction accuracy.

\section{Literature Review}
\label{sec:related}

\subsection{Deep learning and fuzzy learning}

Deep learning \cite{LeCBen15_y} has been successfully applied in many fields, 
such as computer vision and intelligent transportation \cite{LiaDu18}.  
Among all deep learning techniques, Convolutional Neural Networks (CNNs) and Recurrent Neural Networks (RNNs)
are two popular models.
CNNs \cite{LiaWan17} are often deployed to model data with spatial feature interactions 
while RNNs are often used to process data with temporal feature interactions. 
A special RNN called LSTM is widely adopted to overcome the vanishing gradient problem in traditional RNNs \cite{HocBen01_y}.
However, neither CNNs nor LSTMs are perfect models for addressing spatial-temporal problems.
 
To handle imperfect data, fuzzy learning is a powerful tool and shows better performance than deterministic methods \cite{Zad96}. 
Combining fuzzy theory and neural networks can improve complex data representation with probability distribution over cross-layer units \cite{ZheShe17}.
Even though they have been widely applied in control systems \cite{ZheShe17} and portfolio management \cite{DenRen17}, 
no existing work applies fuzzy neural networks in demand prediction. 

\subsection{Passenger demand/traffic flow prediction}

Passenger demand prediction is closely related to traffic flow prediction. 
Both have the same-format of data and are influenced by external factors.
We review both in this section. Traditional approaches to predict future passenger 
demands only consider temporal information, 
such as ARIMA \cite{SheWil08_y,MorGam13_y} or ANN \cite{ChaDil12_y}. 

Recent advances in deep learning \cite{LeCBen15_y} motivate researchers to apply 
deep learning techniques for passenger demand and traffic flow prediction. 
Recent studies employ CNNs to capture complex spatial feature interactions \cite{ZhaZhe17_y} or 
RNNs (including LSTMs) to capture temporal feature interactions \cite{ZhaChe17_y,YuLi17_y}. 
Pioneering works combine CNNs and RNNs to capture both spatial and temporal feature interaction in the data recently. 
Yao et al. \cite{YaoWu18} propose a multi-view model, which employs a CNN and an LSTM  to capture 
the spatial and temporal feature interactions sequentially, but not simultaneously, which can potentially lead to temporal information loss. 
The above works either captures only one of the spatial and temporal feature interactions or captures both sequentially.
None of the methods fully captures spatial-temporal feature interactions simultaneously. 

ConvLSTMs \cite{ShiChe15_y}  are another deep learning model, which combines CNNs and LSTMs.
A ConvLSTM can simultaneously capture the spatial and temporal feature interactions. 
It replaces the fully connected layer in the traditional LSTM with a convolutional layer, 
which shows better performance than the traditional LSTM in precipitation nowcasting.
A follow-up work \cite{KeZhe17} uses ConvLSTMs to predict the passenger demands.
The model is composed of a ConvLSTM and a LSTM to process the weather information, the travel time rate and demand intensity and 
simply fuses the results from the two networks.
However, The model fails to consider the inaccuracy of external data and the inaccuracy of simple prediction result fusion. 
AttConLSTM, a multi-step model built upon the attention-based encoder-decoder framework for passenger demand prediction 
is proposed in \cite{ZhoShe18}. 
However, it fails to consider external factors, which greatly influences passenger demands.
Different from all previous models, our model captures temporal-spatial feature interactions simultaneously without information loss, 
employs fuzzy neural network to handle external data inaccuracy and includes a new and effective feature fusion method based on 
convolutions.

\section{Problem Formulation}
\label{sec:prb}

Being consistent with existing works, we make the following definitions.
Based on the definitions, we present the problem statement of this paper.

\begin{Definition}\rm\textbf{Region}
In this paper, we predict the passenger demands in different areas in a city. The whole city is partitioned into $W\times H$ equal-size grids. 
A grid is called a region, which is denoted by $r$. Let $r_{i,j}$ denote the region with the coordinate $i, j$, where $i\in[0, W)$ and $j\in[0, H)$.
\end{Definition}


\begin{Definition}\rm\textbf{Service request}
A service request $s_k$ made by a passenger is composed of the request ID, pick-up coordinates $s_{k,pc}$ (longitude and latitude), 
and pick-up time $s_{k,pt}$. (Note that we do not consider drop-off location and time for demand predictions.) 
A service request $s_k = \{s_{k,pc}, s_{k,pt}\}$.
A valid request's pick-up location should be in the city.
If it is outside the city, it is discarded.
The total number of available legal requests is denoted by $N$.
\end{Definition}

\begin{Definition}\rm\textbf{Passenger demands}
Time is divided into equal intervals.
The $t$\textsuperscript{th} time interval, starting from 0, is the interval of $\left[t\times C, (t+1)\times C\right)$, 
where $C$ is a constant representing the interval's time span.
The passenger demand of a region $r_{i,j}$ is accumulated in the specific $t$\textsuperscript{th} time interval based on the requests' pick-up time.
The passenger demand in region $r_{i,j}$ at the $t$\textsuperscript{th} time interval 
is denoted by $d_{i,j}^t$.
\begin{equation}
d_{i,j}^t = |\{k\in[0,N): s_{k,pc}\in r_{i,j} \land s_{k,pt}\in [t\times C, (t+1)\times C)\}|.
\end{equation}
The whole area's demands are denoted by $\boldsymbol{D}$, which means
\begin{equation}
\boldsymbol{D}^t = \{d_{i,j}^t|\forall i \in [0, W), \forall j \in [0,H)\}.
\end{equation}
We can imagine that $\boldsymbol{D}^t$ is a demand snapshot of the whole area at the $t$\textsuperscript{th} time interval, 
where every pixel is the demand of that particular location.
\end{Definition}
 
\begin{Definition}\rm\textbf{External information}
Let $\boldsymbol{e}^t$ denote the external information set at the $t$\textsuperscript{th} time interval.
The external factors impacting the passenger demands considered in the paper includes the weather, the day in a week and being daytime or nighttime. 
The process details of the external information will be given in next section.
\end{Definition}

\textbf{Problem statement} 
The problem is defined as follows.
Suppose the current time interval is $t$. 
Given the historical passenger demands and the external information at the $t$\textsuperscript{th} time interval, our goal is to predict the passenger demands in all regions in the city at the $(t+1)$\textsuperscript{th} time interval. 
Specifically, in our problem, we take the historical data, demands and external information 
in the last $k$ time intervals as input and the output is the predicted passenger demands at the $(t+1)$\textsuperscript{th} time interval.
Let $\hat{\boldsymbol{D}}^{t+1}$ denote the predicted passenger demands at the $(t+1)$\textsuperscript{th} time interval.
$\hat{\boldsymbol{D}}^{t+1}$ is a function $f$ of the previous $k$ time intervals' data.
\begin{equation}
	\hat{\boldsymbol{D}}^{t+1} = f(\boldsymbol{D}^{t-k}, \boldsymbol{D}^{t-k+1}, ..., \boldsymbol{D}^{t}, \boldsymbol{e}^{t-k}, \boldsymbol{e}^{t-k+1}, ..., \boldsymbol{e}^{t}).
\end{equation}
Our goal is to minimize the difference between $\hat{\boldsymbol{D}}^{t+1}$ and the true passenger demands $\boldsymbol{D}^{t+1}$. 

\section{STEF-Net}
\label{sec:algorithm}

\begin{figure}
\begin{center}
\includegraphics[width=.4\textwidth]{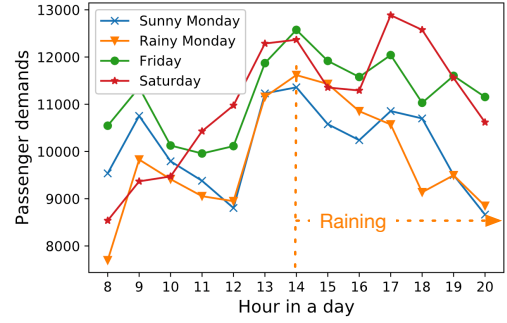}
    \caption{Different hourly demands in different days}
\label{fig:demand}
\vspace{-4mm}
\end{center}
\end{figure}

\subsection{Preliminary analysis} 

In this section, we conduct a preliminary data analysis to provide some intuition on how  
passenger demands are influenced by different factors. 
We use a dataset from Didi Chuxing, China. 
The data contains over 5.24 million service requests from 11/01/2016 to 11/30/2016. 
Fig. \ref{fig:demand} shows the total passenger demands over different hours in different days. 
We pick two Mondays and a Friday as representatives of weekdays and a Saturday as that of weekends.
We select a rainy Monday and a sunny Monday to show the impact of weather on passenger demands. 
Except the rainy Monday, all the other days are sunny.
In the figure, the x axis is the hour of the day and the y axis is the passenger demands during the hour.
We can see the passenger demands have different patterns in different days, at different time of the day, 
and under different weathers. 
For example, at the noon time, the demand drops on Monday and Friday but it increases on Saturday. 
About the weather factor, on the rainy Monday, it starts to rain at 2pm. 
Comparing the sunny Monday with the rainy Monday, we can see that the patterns of passenger demands on the two days before rain are similar while the patterns become different after rain comes.
Specifically, the number of passenger demands keeps decreasing on rainy Monday while there is an increase on sunny Monday. 
The figure shows the passenger demands are determined by complex interactions among many factors.

\begin{figure*}
\begin{center}
\includegraphics[width=.75\textwidth]{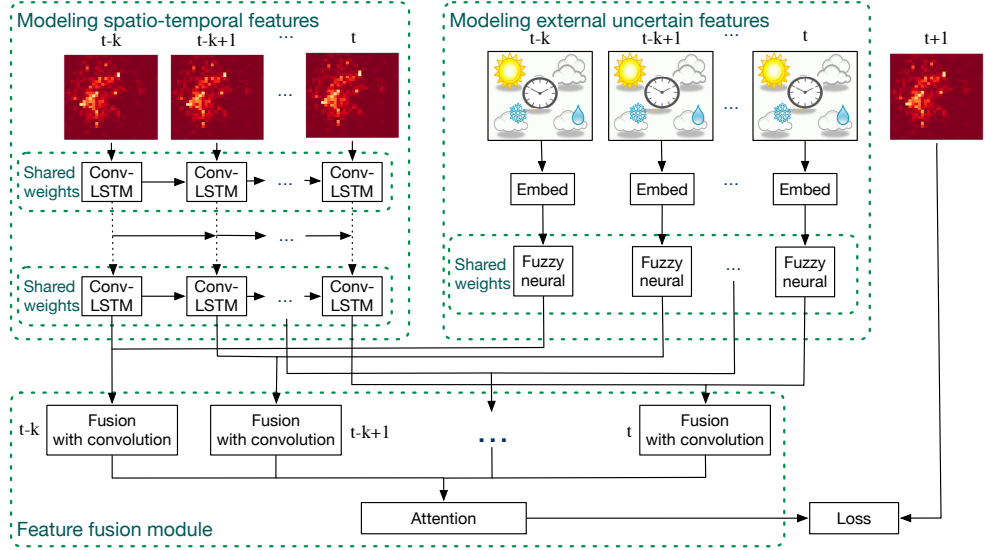}
\caption{Our deep learning model STEF-Net} 
\label{fig:deep_learning}
\vspace{-5mm}
\end{center}
\end{figure*}

\subsection{Overview of our deep learning model }


We propose a deep learning model, STEF-Net, to predict the passenger demands incorporating 
the complex interactions between various factors. 
Our model is illustrated in Fig. \ref{fig:deep_learning}, which is mainly composed of four components. 
(1) As shown in the left side of the figure, we employ a stacking ConvLSTM to capture the \emph{spatial-temporal feature interaction} with the passenger demands. 
The input is the historical passenger demands with location and time information, and the output is the prediction using only spatial-temporal information. 
(2) As shown in the right side of the figure, in parallel, we employ a fuzzy neutral network to capture the 
\emph{external information's interactions} on passenger demands. 
The input is the data about external information and the output is the prediction using external information. 
(3) As shown on the bottom of the figure, the outputs from the stacked ConvLSTMs and the fuzzy neural network
are fused into one network to generate the final output. We propose a new feature fusion method using convolution 
on the data from the same  time period, which keeps the temporal relation of the outputs from the two networks.
Considering the passenger demands are unequally influenced by different time intervals, 
we further propose to adapt an attention layer on the high-level fused data to capture the temporal relevance.
The data are then reshaped into the output format, which matches the regions in a city.
(4) As shown in the right corner of the figure, we employ a loss function to measure the difference between the predicted value and true value. 
A neural network's goal is to minimize the loss defined by an objective function. 
In the following, every component is presented in detail.

\subsection{Modeling spatio-temporal features }

We propose to simultaneously capture the deep spatial and temporal dependencies in passenger demands by stacking ConvLSTMs. 
A ConvLSTM is a neural network model that combines convolutional operation and LSTM units, 
where an LSTM is known to well handle temporal feature interaction without the vanishing gradient problem, 
while convolutional networks are known to gracefully handle spatial feature interactions. 
A ConvLSTM uses the convolution operation to replace the full connections in traditional LSTMs.
In an LSTM,  all the elements are 1D tensors, which accepts the input from $T\times L$ dimensions and generates outputs into $T\times L'$ dimensions.
$T$ is the length of the time sequences, $L$ is the length of one input vector, and $L'$ is the length of one output vector. 
A ConvLSTM transfers all the inputs, memory cell values, hidden states, and various gates in an LSTM into 3D tensors, 
where the first two dimensions are considered as the spatial information, rows and columns and the last dimension is the channels.

To adapt ConvLSTM in our problem, 
we treat the first two dimensions in the passenger demand data as rows and columns at one time interval. 
The ConvLSTM in our problem can be considered as a function $R^{T\times W\times H\times L} \to R^{T\times W\times H\times L'}$, 
where $T$, $L$ and $L'$ are the same as those in the traditional LSTM, 
and $W$ and $H$ are the length of rows and columns, corresponding to the width and height of the grids in the city-wide area in our problem.
Several ConvLSTM layers are stacked in our neural network.
In the last ConvLSTM layer, the length of the output vector is set 1, which means the output is 
a 4D tensor with the size of $T\times W\times H\times 1$. 
This is equivalent to a 3D tensor with the size of $T\times W\times H$.

\subsection{Modeling external uncertain features}

Passenger demands are influenced by many external factors in addition to location and time which are handled in the previous section. 
We need to identify the factors and obtain corresponding data to train our model. 
Note that existing data about external factors are likely inaccurate or of coarse granularity, and thus we propose for the first time to 
employ a fuzzy neural network to model the data.  In the section, we first present our factor selection and data pre-processing, 
and then we present our fuzzy neural network modeling. 
 
\subsubsection{Factor selection and data pre-processing} 

Our model aims to include all highly correlated factors and chooses (1) the weather, (2) the day in a week, and (3) being daytime or nighttime.
Obviously, weather greatly influences a passenger's choice between taking uber and walking, waiting and then taking a bus.
When it rains heavily or when it is very cold or very hot, people tend to take a more comfortable way of transportation.  
In our model, the weather is represented by temperature, dew point, humidity, pressure, wind speed, and weather condition. 
The first five variables are numerical variables. 
The last one, weather condition, is represented by ten different categories: clear, partly cloudy, scattered cloud, mostly cloudy, haze, light rain, shower, mist, patches of fog, and fog. The categories are indexed from 1 to 10.
The numbers are embed using one-hot vectors with 10 dummy variables.
The second factor, the day in a week, refers to the seven days in a week,
which also affluences people's daily transportation behavior. 
This factor is a categorical variable and is represented by a one-hot vector with 7 dummy variables. 
The third factor is whether it is dark outside and it is differentiated by the sunrise and sunset time of the day.
In summary, to represent all the three factors, we choose 24 variables in total in our model. 
The external information is 1D tensor.

\subsubsection{Fuzzy neural network modeling } 

\begin{figure}
\begin{center}
\includegraphics[width=.4\textwidth]{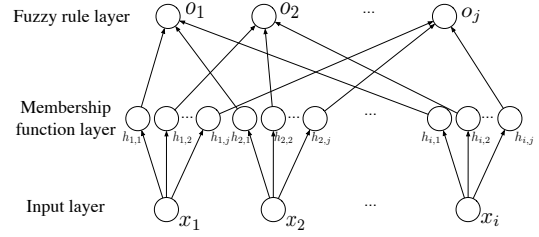}
    \caption{Illustration of the fuzzy neural network}
\label{fig:fuzzy}
\vspace{-6mm}
\end{center}
\end{figure}

We adopt a fuzzy neural network to learn the representation of the external information. 
The fuzzy neural network is composed of two hidden layers, membership function layer and logic rule layer.
The architecture is shown in Fig. \ref{fig:fuzzy}.
The membership function layer calculates the degree that an input node belongs to a certain fuzzy set.
Let $x_i$ denote the $i$\textsuperscript{th} element in the input.
In the membership function layer, every element is split by multiple Gaussian distributions.
Let $j$ index the Gaussian distribution for the $i$\textsuperscript{th} element. One distribution is denoted by ($\mu_{i,j}$, $\delta_{i,j}$).
The membership function layer's output is calculated as follows,
\begin{equation}
    h_{i,j} = e^{\frac{-(x_i-\mu_{i,j})^2}{\delta_{i,j}^2}}.
\end{equation}
The logic rule layer performs the ``AND'' fuzzy logic operation as follows,
\begin{equation}
    o_{j} = \prod_{i} h_{i,j}. 
\end{equation}
Through the rule layer, the output can present the probability that it is related to every unit.
In the fuzzy neural network, all time intervals' external information shares the same member function layer and logic rule layer.

The output from the fuzzy neural network of one interval's external information is reshaped into two dimensions to match the passenger demands in a whole city, which is $W\times H$.
The outputs from the past $T$ time intervals' external information are a 3D tensor with the shape size of $T\times W\times H$.

\subsection{Feature fusion module}

In the above two sections, the passenger demand data and external information are mapped 
into the same feature representation format from two separate networks.
They need to be combined to predict the next time interval's demand.

Previous works employ weighted addition \cite{KeZhe17,ZhaZhe17_y} to fuse the two components. 
In this paper, we propose to use convolutional operation to fuse the outputs from two networks. 
In addition, to keep the temporal feature interaction, we first fuse the data from the same time interval and then apply an attention layer to generate the final output.

\begin{figure}
\centering     
\subfigure{\label{fig:con}\includegraphics[width=0.4\textwidth]{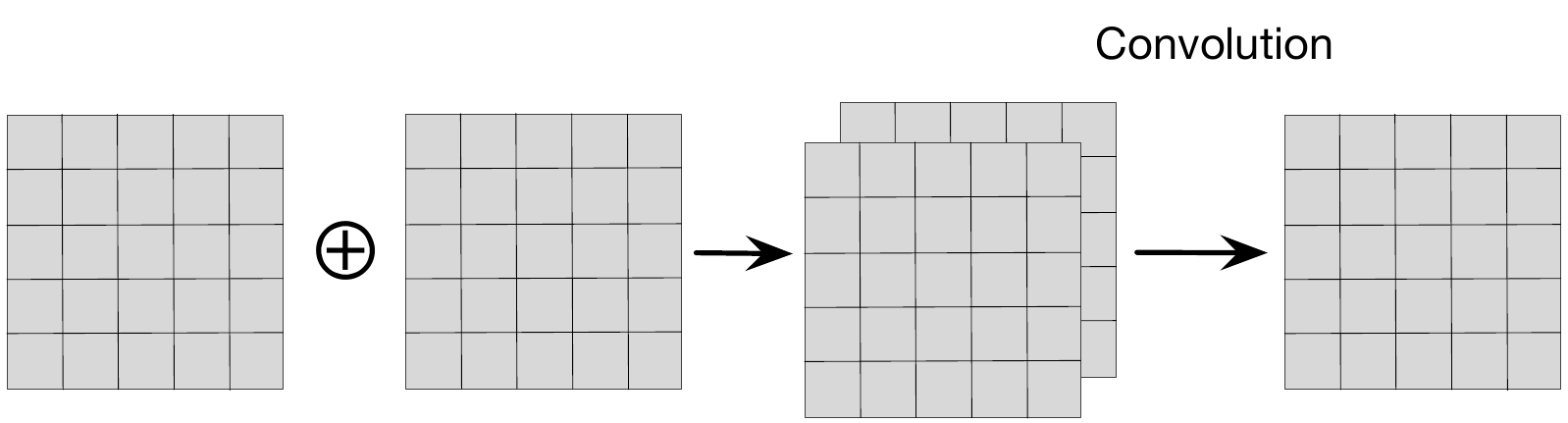}}
    \caption{The fusion method with convolution}
\label{fig:fuse}
\end{figure}

The outputs of the passenger demand and external information at the $t$\textsuperscript{th} are denoted by $\mathbf{O}_{t,p}$ and $\mathbf{O}_{t,e}$, respectively.
Let $\mathbf{O}_{t,f}$ denote the output after fusion with convolution denoted by $\oplus$.
The illustration of fusion with convolution is shown in Fig. \ref{fig:fuse}.
The calculation can be presented by the following equation,
\begin{equation}
    \mathbf{O}_{t,f} = \mathbf{O}_{t,p}\oplus \mathbf{O}_{t,e}.
\end{equation}
We concatenate the two by adding a new dimension, which can be imaged as the channel in a CNN.
After concatenation, $\mathbf{O}_{t,f}\in R^{W\times H\times2}$.
To make the output's dimension consistent, we apply a convolutional operation with window size $w\times h$ ($w \ll W \& h \ll H$) with 1 channel, which outputs a 2D tensor with the size of $W\times H$.
In this way, our fusion with convolution method only needs $w\times h$ parameters while the weighted addition requires $2\times W\times H$ parameters in previous works \cite{KeZhe17,ZhaZhe17_y}.
In addition, the convolutional operation can further learn the spatial information on the fused data.

After fusing the data, we use a bidirectional LSTM and attention to further capture the temporal relevance.
In the bidirectional LSTM,  the data are flattened into one dimension and fed into LSTMs:
\begin{eqnarray}
    \overrightarrow{\mathbf{h_t}} =& \overrightarrow{LSTM}\left(\mathbf{w_t}, \overrightarrow{\mathbf{h_{t-1}}}\right),\\
    \overleftarrow{\mathbf{h_t}} =& \overleftarrow{LSTM}\left(\mathbf{w'_t}, \overleftarrow{\mathbf{h_{t+1}}}\right).
\end{eqnarray}
$w_t$ and $w'_t$ are the weights in the forward and backward LSTMs, respectively.
The outputs from the forward and backward are added in an element-wise way:
\begin{equation}
    \mathbf{h_t} = \overrightarrow{\mathbf{h_t}}+ \overleftarrow{\mathbf{h_t}}.
\end{equation}
We keep the number of units in the LSTM being same as the number of grids, which is equal to $W\times H$.
All the time steps are concatenated into a matrix H:
\begin{equation}
\begin{split}
    \mathbf{H} = (\mathbf{h_1}, \mathbf{h_2}, ..., \mathbf{h_t}).
    \end{split}
\end{equation}
For all time steps' values in every grid, we use a linear transformation and a \textit{softmax} activation function to get the attention weights on the time step domain:
\begin{equation}
\begin{split}
    \mathbf{a} = \mathrm{softmax}(\mathbf{WH}).
    \end{split}
\end{equation}
The outputs are the weighted sum of the hidden states and the attention weights in an element-wise way.
The passenger prediction is the outputs after being weighted by attention.

\subsection{Objective function}
In our model, the objective function is defined as the mean squared error between the true passenger demands and predicted passenger demands.
The model is trained based on mini-batches.
Suppose there are $m$ samples in a mini-batch and every sample is indexed by $i$,
the objective function $L(\mathbf{\theta})$ with trainable parameters $\mathbf{\theta}$ is defined as follows,
\begin{equation}
\label{equ:cost}
	L(\mathbf{\theta}) = \frac{1}{m} \sum_{i=1}^{m} \|\hat{\boldsymbol{D}}^{(t+1)}_i-\boldsymbol{D}^{(t+1)}_i\|^2, 
\end{equation}
It is the mean squared error between the predicted and true passenger demands in a mini-batch.
The optimization algorithm in our model is the ADAptive Moment estimation (Adam) \cite{KinBa14}, which adaptively changes the effective  learning rate  during training. 

\section{Evaluation}
\label{sec:evaluation}


\subsection{Evaluation objectives and metrics}

We evaluate STEF-Net by comparing it with state-of-the-art models on real data
with regarding to the accuracy in passenger demands prediction. 
The accuracy is measured by two metrics, Mean Absolute Error (MAE) and Rooted Mean Squared Error (RMSE). 
The MAE and RMSE are two widely employed metrics to evaluate the performance of a prediction system \cite{ZhaZhe17_y}.


\subsection{Dataset}

The dataset used for training and testing is from Didi Chuxing. 
The data contains over 5.24 million non-duplicating service requests from 11/01/2016 to 11/30/2016 in Chengdu City, China. 
In the dataset, every service request record 
is composed of the request ID, pick-up time, pick-up coordinates, drop-off time and drop-off coordinates. 
(Note that we do not need drop-off time and drop-off coordinates here.) 
The data of the first 23 days is used for training and that of the last 7 days (one week) for testing.
The area is about 14.41km $\times$ 14.39km.
We divide the whole area into 20$\times$20 same-size grids.
The length and width of every grid are both about 700 meters.
The time interval is set half an hour.
Same as previous studies \cite{ZhaZhe17_y,ZhoShe18}, the passenger demands are scaled into [0,1] using max-min scaling. 
In the final step, the demand values are recovered by the inverse of max-min scaling. 

Regarding the external information, the day of a week is extracted from the pick-up time.
The weather information and sunrise/sunset information are crawled from the Weather Underground website \cite{WG18} using Python. 
The website provides historical weather information in Chengdu.
As presented in previous section, we extract 24 features about the external information. 


\subsection{Hyperparameters and development environment}

In our network, 3 ConvLSTMs are stacked and all the ConvLSTMs use 64 filters of size 3$\times$3.
A convolutional layer with one filter is added after the stacked ConvLSTMs to convert the data into a 20$\times$20 tensor.
The membership layer in the fuzzy neural network is composed of 24$\times$400 units, and the fuzzy rule layer has 400 units.
The outputs from the fuzzy neural network are reshaped into 20$\times$20 and the following convolutional layer has one filter with the kernel size of 3$\times$3.  
In the fusion part, the convolutional layer has one filter with the kernel size of 3$\times$3.  
In the fully connected layers, there are three layers with 100 units, 200 units, and 400 units, respectively.
The output is reshaped into 20$\times$20 to match the prediction in the whole city.
The parameters in the fuzzy neural network are uniformly initialized from 0 to 1. 
All the other parameters are uniformly initialized.

In our model, to be consistent with the previous studies we compare with \cite{KeZhe17,ZhoShe18}, 
we by default use previous 8 time intervals' data to predict the current time interval's passenger demands, 
It means the current time interval's passenger demands are predicted based on the historical 4 hours' data.
We also use 2 hours' data to predict the demands to see how the time length can influence the prediction accuracy. 

Our model is developed on the top of Keras \cite{Cho15} with the backend of Tensorflow \cite{AbaAga16}.
The model is running on a desktop with an Intel Xeon 3.10GHz$\times$4 CPU and a GeForce GTX 1050 Ti GPU.
The model is trained by mini-batches. 
Every mini-batch has 16 samples.
Every model is trained using 50 epochs and the results are generated after that.

\subsection{Baselines}
\label{subsec:bl}

For a thorough comparison with existing methods, 
we compare our model with three categories of methods, time-series (ARIMA), regression-based (Ridge and XGBoost) and neural network-based methods (ST-ResNet, AttConLSTM and FCL-Net).
The methods are presented as follows: 
\begin{itemize}
\setlength\itemsep{-0.2em}
    \item \textbf{ARIMA \cite{MorGam13_y}:} ARIMA uses both moving average and autoregressive to predict the next time interval's passenger demands.

    \item \textbf{Ridge linear regression \cite{HoeKen70}:} Ridge linear regression uses a linear equation to model 
    the relationship between historical features and future passenger demand. 
    We reshape all features in this paper to a vector and feed vectors into the linear regression.
    \item \textbf{XGBoost \cite{CheGue16}:} XGBoost (2016) is a widely used boosting method with a tree structure. 
    All features are also reshaped into vectors to feed the XGBoost model.
\item \textbf{AttConLSTM\cite{ZhoShe18}:} AttConLSTM (2018) fuses attention and ConvLSTM into an auto-encoder model.
It stacks CNNs and ConvLSTMs to encode and decode the passenger demands and extracts passenger demands patterns as references in an attention network.
\item \textbf{FCL-Net \cite{KeZhe17}:} FCL-Net (2017) employs ConvLSTM and LSTM to extract information from demands, 
time and weather. It fuses the outputs from two networks by addition. 
\item \textbf{ST-ResNet \cite{ZhaZhe17_y}:} 
ST-ResNet (2017) uses ResNet to capture the spatial and temporal information on demands 
from three categories, recent, near and distant. Only weather information at the current time interval is considered.
It fuses data by addition.
Because the distant demands require at least three weeks for one sample, we only take the recent and near categories.
\end{itemize}
All the models follow the settings in their original papers, and all of them are trained 50 epochs. 

\subsection{Results}

\subsubsection{Comparisons with baselines}
We present the comparison results with baselines in Table \ref{table:rst}.
From this table, we can see our model outperforms all the others regarding to both metrics. 
When the historical data is 4 hours, our model can achieve 3.89 in RMSE and 2.27 in MAE.
The results at least 9.9\% better in MAE and 11.3\% better in RMSE than the best one among all baseline methods 
and about 34.6\% better in MAE and 37.8\% better in RMSE than the worst among all baselines. 
Ridge linear regression performs worst because it only considers the linear relation among features.
Note that after the model is trained, we use 2 hour data and 4 hour data to predict future demands, respectively. 
When more historical data is utilized to predict, the performance improves.

\begin{table}
\centering
\label{table:rst}
    \begin{tabular}{lcccc}
\hline
\hline
	    \multirow{2}{*}{Model name} & \multicolumn{2}{c}{2 hours} & \multicolumn{2}{c}{4 hours} \\
	    \cline{2-5}
	    & MAE & {RMSE}  & MAE & {RMSE} \\
\hline
        ARIMA & 3.61 & 6.42 & 2.85 &4.91\\ 
        Ridge linear &3.50&6.32& 3.47& 6.25\\
        XGBoost &3.48&6.18 & 3.29 &5.87\\
        ST-ResNet  &2.90& 5.15 &2.86 &5.02 \\
        AttConLSTM &2.63 & 4.58 &2.60 &4.55 \\
        FCL-Net &2.58& 4.46 & 2.52 & 4.39\\
        STEF-Net &2.31&4.05&\textbf{2.27}&\textbf{3.89}\\
\hline
\hline
\end{tabular}
\vspace{-0.5cm}
\end{table}

\begin{figure}[h!]
\centering     
\hspace*{\fill}%
\subfigure[The second last time step in a sunny day]{\label{fig:c1}\includegraphics[width=0.23\textwidth]{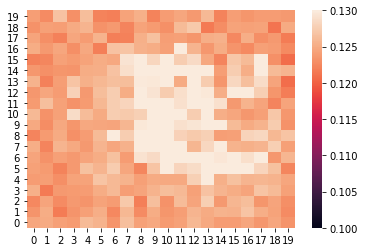}}
\hfill
\subfigure[The last time step in a sunny day]{\label{fig:c2}\includegraphics[width=0.23\textwidth]{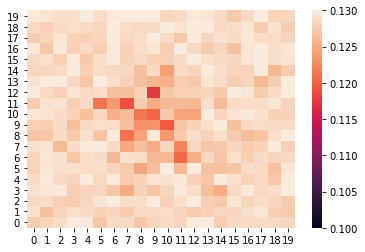}}
\subfigure[The second last time step in a rainy day]{\label{fig:r1}\includegraphics[width=0.23\textwidth]{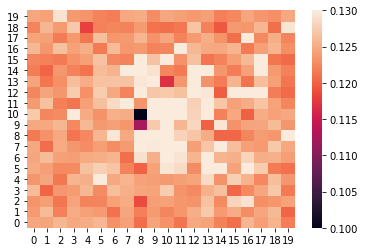}}
\hfill
\subfigure[The last time step in a rainy day]{\label{fig:r2}\includegraphics[width=0.23\textwidth]{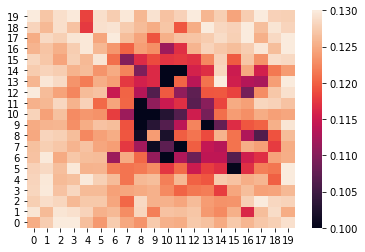}}
    \caption{The attention probabilities of the historical time steps that influence the current passenger demands under different weather conditions}
\label{fig:h1}
\end{figure}

\subsubsection{Qualitative results}
In this section, we illustrate how the passenger demands are influenced by the historical data.
Using heatmaps, we plot the attention weights under different weather conditions (sunny or rainy) at 8pm on two different days in Fig. \ref{fig:h1}.
In each day, we plot the attention weights of the last time step and the second to the last time step, which represent the intermediate history and 
next intermediate history.
Note that there are 20$\times$20 grids, and thus in the heatmaps, every small square represent one geographic grid. 
Center grids represent the downtown area of the city and border grids represent suburban areas. 
Comparing Fig. \ref{fig:c1} with Fig. \ref{fig:c2}, and comparing Fig. \ref{fig:r1} with Fig. \ref{fig:r1}, 
we can see that the current passenger demands in the city center are more influenced by the second to the last time step 
than by the last one while those in the border area are more influenced by the last time step regardless of the weather.
Comparing figures in the sunny day and rainy day in Fig. \ref{fig:h1}, we can also see that the weather greatly influences the passenger demands.
When the outside is rainy, the passenger demands are less influenced by the last time step in the center of the city (downtown).
The rainy weather may incur passengers in the downtown area to change their travel plan 
by pre-scheduling their activities, e.g., reducing the outdoor activities,
while the users in the suburbs are not influenced much.
%

\subsubsection{Ablation studies: comparisons with variants of STEF-Net}

In our evaluation, we also compare STEF-Net with its variants to explore how different components influence the prediction performance.
We explored the following variants, 
\begin{itemize}
\setlength\itemsep{-0.2em}
\item \textbf{ConvLSTM $\rightarrow$ CNN\&LSTM :} 3 convolutional layers with the 64 windows of size $3\times 3$ and 3 LSTM layers with 64 units in each layer are stacked to replace ConvLSTMs for the  demand information processing.
\item \textbf{ConvLSTM $\rightarrow$ LSTM:} 3 LSTM layers with 64 units in each layer are stacked to replace ConvLSTMs for the demand  information processing.
\item \textbf{Fuzzy $\rightarrow$ LSTM:} 3 LSTM layers with 64 units in each layer are stacked to replace the fuzzy neural network for the external information processing.
\item \textbf{No attention layers:} no attention layers are used after the outputs from the ConvLSTM and the fuzzy neural network are fused.
\item \textbf{No external information:} no external information is used in the model. Features from the ConvLSTM are directly fed to the attention layer.
\item \textbf{Weighted addition:} weighted addition is used to replace our convolutional operation in data fusion.
\end{itemize}

\begin{table}
\centering
\label{table:rst-df}
    \begin{tabular}{lcc}
\hline
\hline
        {Model name} & MAE & {RMSE} \\
\hline

        ConvLSTM $\rightarrow$ CNN\&LSTM&2.32 &4.01\\
        ConvLSTM $\rightarrow$ LSTM& 2.36 & 4.09\\
        Fuzzy $\rightarrow$ LSTM& 2.58 & 4.51 \\
        No attention layers &2.59& 4.35\\
        No external data & 2.40 & 4.24\\
	Weighted addition &2.38 &4.09\\
        STEF-Net &\textbf{2.27}&\textbf{3.89}\\
\hline
\hline
\vspace{-0.6cm}
\end{tabular}
\end{table}

The results are shown in Table \ref{table:rst-df}.
We can see that our model, STEF-Net, has the best performance among all variants.
Specifically, our model can achieve the smallest RMSE and MAE compared to its variants.
Comparing to these variants, we can see that the ConvLSTM can effectively capture the spatio-temporal information, which performs better than both CNN\&LSTM and LSTM, and the fuzzy neural network can outperform the LSTM in processing the external information.
In addition, we can see that attention layers can further capture the temporal relevance and improve the performance. 
Among all the components, fuzzy neural network can significantly improve the performance comparing to the LSTM.
The feature fusion method can further capture the temporal feature interaction and the weighted addition performs worse than the fusion with convolution.
Comparing the results between our model with the model withoout external information, we can see that the external information is important in predicting the passenger demands, which matches our intuition of fusing it into our model. 

\section{Conclusion}
\label{sec:con}
In this paper, we propose a Spatio-TEmporal Fuzzy neural Network (STEF-Net) to accurately predict passenger demands in the near future.
Our model can effectively capture complex input dependencies, including spatial, temporal and external factors, which may influence 
future passenger demands. In the proposed approach, we combine deep learning with a fuzzy neural network to model spatio-temporal and external information, respectively.
We employ a new feature fusion method with convolution followed by an attention layer, to fuse two neural networks into one and keep temporal relations for 
further temporal relevance modeling. 
Extensive experiments on real-world dataset show that, our model outperforms the state-of-the-art approaches with over 10\% improvement in RMSE.

\bibliographystyle{IEEEtran}
\bibliography{ref}
\end{document}